\documentclass[runningheads]{llncs}

\usepackage{algorithm}
\usepackage{algorithmic}
\usepackage{graphicx}
\usepackage[subpreambles=true]{standalone}
\usepackage{subcaption}
\usepackage{float}
\usepackage{pdflscape}
\usepackage{booktabs}
\usepackage{hyperref}

\begin{document}

\title{LUCID--GAN: Conditional Generative Models \\ to Locate Unfairness\thanks{We thank Arne Vanhoyweghen, Brecht Verbeken, and Bert Verbruggen for stimulating discussions and feedback on earlier drafts of this work. This project benefited from financial support from Innoviris. Any remaining errors or shortcomings are those of the authors.}}
\author{Andres Algaba\inst{1} \and
Carmen Mazijn\inst{1} \and
Carina Prunkl\inst{2} \and
Jan Danckaert\inst{1} \and
Vincent Ginis\inst{1,3}}
\authorrunning{A. Algaba et al.}
\institute{Vrije Universiteit Brussel, Pleinlaan 2, 1000 Brussels, Belgium 
\email{\{andres.algaba,carmen.mazijn,jan.danckaert,vincent.ginis\}@vub.be} \and
Oxford University, Wellington Square, OX1 2JD, Oxford, United Kingdom 
\email{carina.prunkl@philosophy.ox.ac.uk} \and Harvard University, 150 Western Ave, Boston, USA}

\maketitle

\setcounter{footnote}{0}

\begin{abstract}
 Most group fairness notions detect unethical biases by computing statistical parity metrics on a model's output. However, this approach suffers from several shortcomings, such as philosophical disagreement, mutual incompatibility, and lack of interpretability. These shortcomings have spurred the research on complementary bias detection methods that offer additional transparency into the sources of discrimination and are agnostic towards an a priori decision on the definition of fairness and choice of protected features. A recent proposal in this direction is LUCID (Locating Unfairness through Canonical Inverse Design), where canonical sets are generated by performing gradient descent on the input space, revealing a model's desired input given a preferred output. This information about the model's mechanisms, i.e., which feature values are essential to obtain specific outputs, allows exposing potential unethical biases in its internal logic. Here, we present LUCID--GAN, which generates canonical inputs via a conditional generative model instead of gradient--based inverse design. LUCID--GAN has several benefits, including that it applies to non--differentiable models, ensures that canonical sets consist of realistic inputs, and allows to assess proxy and intersectional discrimination. We empirically evaluate LUCID--GAN on the UCI Adult and COMPAS data sets and show that it allows for detecting unethical biases in black--box models without requiring access to the training data.\footnote{Code is available at \url{https://github.com/Integrated-Intelligence-Lab/canonical_sets}.}

\keywords{algorithmic fairness \and bias detection \and discrimination \and \\ generative models.}
\end{abstract}

\section{Introduction}
\label{sec:intro}
The increasing use of Artificial Intelligence (AI) algorithms in (semi--)automated decision--making processes has raised concerns about harmful and discriminatory decision patterns observed in contexts such as healthcare \cite{ledford_millions_2019,obermeyer_racialbias_2019}, education \cite{makhlouf_fairness_2021}, and hiring \cite{chen_gender_2018,raghavan_hiring_2020}. In these cases, the algorithmic decisions discriminate against people based on (legally) protected features, including gender, age, and ethnicity \cite{lindholm_discrimination_2022}. Often, the very detection of discrimination is difficult because protected characteristics are encoded in so--called proxies. Proxy discrimination is especially prevalent in the era of big data, where algorithms can reconstruct many protected features from non--protected data \cite{barocas_law_2016}. In this paper, we address this challenge by developing a fairness evaluation method that reveals an algorithm's desired feature values for a given outcome. Our method exposes proxies that embed potential unethical biases and enhances transparency in the algorithm's decision--making process.

Algorithmic discrimination can be direct or indirect \cite{barocas_fairml_2019}. Indirect discrimination focuses on the impact of a given decision on a protected group. Within US law, this is often labeled ``disparate impact.'' While algorithmic decision--making tools have been shown to frequently put members of particular social groups at a disadvantage \cite{barocas_fairml_2019}, indirect discrimination can often be justified as being a proportionate means of achieving a legitimate goal (e.g., in hiring decisions) \cite{adams_2023_proxy}. Direct discrimination, on the other hand, focuses not so much on the impacts but on the \textit{reasons} behind a given decision. In other words, a person not being hired because of their belonging to a particular social group would constitute a case of direct discrimination.  While direct discrimination is illegal both under EU and US law\footnote{Anti-discrimination law in the US speaks of ``disparate treatment'' which, while similar to direct discrimination, additionally requires there to be discriminatory intent \cite{prince_proxy_2019}.}, cases of direct discrimination often go unchallenged due to the difficulty of establishing a causal link between protected characteristics and decision outcomes. Within the context of algorithmic decision making, the direct discrimination doctrine is often translated into the requirement to abstain from using protected characteristics as input variables. In practice, however, such attempts of ``fairness--through--unawareness'' rarely work since protected features are often encoded through other features, giving rise to potential proxy discrimination \cite{fazelpour2021algorithmic}.

Such proxy discrimination further exacerbates the challenge of establishing potential causal links between protected characteristics and less favorable decision outcomes. The identification of proxies and their relationship to protected characteristics is therefore crucial to identify cases of directly discriminating algorithms \cite{adams_2023_proxy}. 

While proxy discrimination plays an important role for direct discrimination, it is more often considered in the context of indirect discrimination by algorithms \cite{barocas_fairml_2019,wachter_2021_automate}. Indirect discrimination has a strong focus on the \textit{outcome} of a given decision (as opposed to the \textit{reason} behind it) and so it more readily connects to the fairness literature's large focus on algorithmic outputs. 
%The literature on algorithmic fairness has primarily focused on defining unfairness and eliminating unethical biases, while recent efforts focus on rigorously detecting it \cite{lum_2022_debias}. 
The traditional approach involves translating philosophical or political notions of group fairness into a statistical parity metric on the model's output \cite{makhlouf_fairness_2021}.\footnote{Note that we focus on notions of group fairness which are most commonly used in practice, while many other definitions of fairness exist, including individual and counterfactual fairness \cite{dwork_2012_fairness,kusner_2017_counterfactual}.} However, output--based fairness evaluations of this kind have several shortcomings. First, many notions of group fairness are incompatible, except for highly constrained special cases \cite{kleinberg_inherent_2016}. Second, the problem of many fairness notions only worsens as there is often substantial philosophical disagreement on which ones are genuinely fair in each specific context \cite{binns_fairness_2018}. Third, by reducing the evaluation to a single number, output--based metrics make it hard to verify the validity of the results and to understand exactly why the model is unfair \cite{meng_mimic_2022}. Fourth, most output--based fairness evaluations make it difficult to detect intersectional discrimination as they are limited to a specific selection of protected features. This selection entails the risk of missing discrimination against people at the intersection of different protected features \cite{buolamwini_gender_2018,kong_intersectionally_2022} or against groups that do not share a protected feature \cite{binns_apparent_2020}. Finally, the computation of these parity metrics often depends on a benchmark data set that may be biased or unbalanced to some extent \cite{northcutt_pervasive_2021,lum_2022_debias}. These shortcomings motivate the research on complementary fairness evaluation methods, which offer additional transparency into the sources of discrimination and are agnostic towards an a priori decision on the choice of protected features, as this is often case-dependent and policy--related \cite{goethals_2022_precof}.

In this paper, we build on the LUCID (Locating Unfairness through Canonical Inverse Design) method proposed by Mazijn et al. \cite{mazijn_sets_2022}. LUCID generates canonical inputs by performing gradient descent on the input space, revealing a trained model’s desired input given a preferred output. The resulting canonical set contains valuable information about the model’s mechanisms, i.e., which feature values are essential to obtain specific outputs. This allows us to expose potential unethical biases in its internal logic by inspecting the distribution of the protected features. Despite LUCID's appealing properties as a fairness evaluation method, the canonical sets generated by gradient--based inverse design have some critical shortcomings. First, while the canonical sets in their current form are specifically suitable for tabular data, they require differentiable models. LUCID thereby omits the class of tree--based models, which are very effective for tabular data \cite{shwartz_deep_2022}. Additionally, the current gradient--based approach may lead to non--realistic canonical inputs, and it is not straightforward to assess proxy or intersectional discrimination \cite{crenshaw_1991_intersection}.

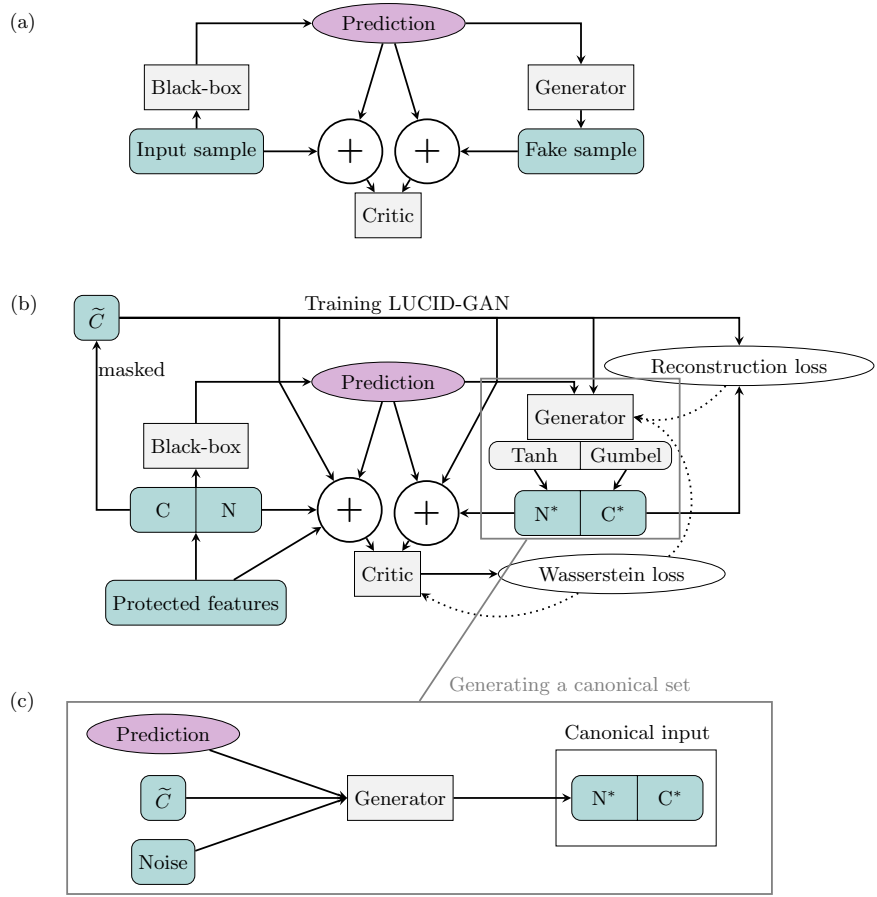
\begin{figure}[H]
  \centering
  \begin{tikzpicture}

% Figure 1
\node (real) [data] {Input sample};
%\node (fair) [data, below of=real, yshift=-0.45cm] {protected features};
\node (blackbox) [model, above of=real] {Black-box};
\node (pred) [prediction, above of=blackbox, xshift=3cm] {Prediction};
\node (generator) [model, below of=pred, xshift=3cm] {Generator};
\node (fake) [data, below of=generator] {Fake sample};
\node (concatreal) [concat, right of=real, xshift=1.4cm]  {\Large \textbf{+}} ;
\node (concatfake) [concat, left of=fake, xshift=-1.4cm] {\Large \textbf{+}};
\node (critic) [model, below of=concatreal, xshift=0.59cm] {Critic};
%\node (text) [rectangle, right of= critic, minimum width=1cm, minimum height=0.8cm,text centered, xshift=2cm] {fake / real ?};

\draw [arrow] (real) -- (blackbox); %  node[anchor=west] {no}
\draw [arrow] (blackbox) |- (pred);
\draw [arrow] (pred) -| (generator);
\draw [arrow] (generator) -- (fake);
\draw [arrow] (pred) -- (concatreal);
\draw [arrow] (pred) -- (concatfake);
\draw [arrow] (real) -- (concatreal);
%\draw [arrow] (fair) -- (concatreal);
\draw [arrow] (fake) -- (concatfake);
\draw [arrow] (concatreal) -- (critic);
\draw [arrow] (concatfake) -- (critic);
%\draw [arrow] (critic) -- (text);
%\draw [arrow] (fair) -- (real);

% \node [rectangle, above of=pred, xshift=0.3cm, yshift=-0.4cm] {LUCID-GAN overview};

% Figure 2
\node (real_2) [splitdata, below of=critic, text width=0.8cm, xshift=-3cm, yshift=-3.6cm] {\nodepart{one} C \nodepart{two} N};
\node (fair_2) [data, below of=real_2, yshift=-0.45cm] {Protected features};
\node (blackbox_2) [model, above of=real_2] {Black-box};
\node (masked_2) [data, above of=blackbox_2, xshift=-1.55cm, yshift=1cm] {$\widetilde{C}$};
\node (pred_2) [prediction, above of=blackbox_2, xshift=3cm] {Prediction};
\node (generator_2) [model, below of=pred_2, xshift=3cm, yshift=0.45cm] {Generator};
\node (tanhsoft_2) [splitmodel, below of=generator_2, text width=1.2cm, xshift=0cm, yshift=0.4cm] {\nodepart{one} Tanh \nodepart{two} Gumbel};
\node (fake_2) [splitdata, below of=generator_2, text width=0.8cm, yshift=-0.5cm] {\nodepart{one} N$^*$ \nodepart{two} C$^*$};
\node (concatreal_2) [concat, right of=real_2, xshift=1.4cm] {\Large \textbf{+}};
\node (concatfake_2) [concat, left of=fake_2, xshift=-1.4cm] {\Large \textbf{+}};
\node (critic_2) [model, below of=concatreal_2, xshift=0.59cm] {Critic};
\node (loss_2) [loss, right of= critic_2, xshift=2.5cm] {Wasserstein loss};
\node (recon_2) [loss, right of= generator_2, yshift=0.8cm, xshift=1.48cm] {Reconstruction loss};

\draw [arrow] (real_2) -- (blackbox_2);
\draw [arrow] (fair_2) -- (real_2);
\draw [arrow] (blackbox_2) |- (pred_2);
\draw [arrow] (pred_2) -| (generator_2.105);
% \draw [arrow] (generator_2) -- (fake_2);
\draw [arrow] (tanhsoft_2.one south) -- (fake_2.one north);
\draw [arrow] (tanhsoft_2.two south) -- (fake_2.two north);
\draw [arrow] (pred_2) -- (concatreal_2);
\draw [arrow] (pred_2) -- (concatfake_2);
\draw [arrow] (real_2) -- (concatreal_2);
\draw [arrow] (fair_2) -- (concatreal_2);
\draw [arrow] (fake_2) -- (concatfake_2);
\draw [arrow] (concatreal_2) -- (critic_2);
\draw [arrow] (concatfake_2) -- (critic_2);
\draw [arrow] (critic_2) -- (loss_2);

\draw [dot, bend left, thick] (loss_2) edge (critic_2);
\draw [dot, bend right=65, thick] (loss_2) edge (generator_2.east);
\draw [dot, bend left, thick] (recon_2) edge (generator_2.east);

\draw [arrow] (real_2.west) -| node[anchor=west, yshift=2.2cm, xshift=-0.1cm] {masked} (masked_2);

\draw [arrow] (masked_2.east) -| (generator_2.60);
\draw [arrow] (masked_2.east) -| (recon_2);
\draw [arrow] (masked_2.east) -| ($(pred_2.west) - (5mm,0)$) -- (concatreal_2);
\draw [arrow] (masked_2.east) -| ($(pred_2.east) - (-5mm,0)$) -- (concatfake_2);
\draw [arrow] (fake_2.east) -| (recon_2);

\node (rec1) [rectangle, thick, right of=pred_2, xshift=2cm, minimum width=3.1cm, minimum height=2.5cm, yshift=-1.2cm, draw=gray] {};

\node [rectangle, above of=pred_2, xshift=0.3cm, yshift=0.2cm] {Training LUCID-GAN};

% Figure 3
\node (pred_3) [prediction, below of= critic_2, xshift=-3.5cm, yshift=-1.5cm] {Prediction};
\node (mask_3) [data, below of=pred_3] {$\widetilde{C}$};
\node (noise_3) [data, below of=mask_3] {Noise};
\node (generator_3) [model, right of=mask_3, xshift=2.7cm] {Generator};
\node (fake_3) [splitdata, right of=generator_3, xshift=2.7cm, text width=0.8cm] {\nodepart{one} N$^*$ \nodepart{two} C$^*$};

\node (rec2) [rectangle, thick, right of=mask_3, xshift=3cm, minimum width=11cm, minimum height=3cm, yshift=0.0cm, draw=gray] {};
% \node [above of=generator_3, yshift=0.5cm] {};

\node [rectangle, right of=generator_3, draw=black, xshift=2.68cm, minimum width=2.5cm, minimum height=1.5cm, yshift=0.0cm] {};
\node [rectangle, above of=fake_3] {Canonical input};

\node [rectangle, above of=generator_3, text=gray, xshift=2.65cm, yshift=0.75cm] {Generating a canonical set};

\draw [thick, draw=gray] (rec1) -- (rec2.north);

\draw [arrow] (pred_3) -- (generator_3.west);
\draw [arrow] (mask_3) -- (generator_3.west);
\draw [arrow] (noise_3) -- (generator_3.west);

\draw [arrow] (generator_3.east) -- (fake_3.west);

% Annotation
\node [left of=pred, text width=0.8cm, xshift=-4.52cm] {(a)};
\node [below of=critic, text width=0.8cm, xshift=-5.5cm, yshift=-0.4cm] {(b)};
\node [below of=critic_2, text width=0.8cm, xshift=-5.5cm, yshift=-1cm] {(c)};

\end{tikzpicture}
  \caption{The LUCID-GAN architecture. (a) High--level overview, (b) LUCID--GAN training mechanism, and (c) the details of the Generator. The input sample may consist of both numerical ($N$) and one--hot encoded ($C$) features, which may contain protected features that are not necessarily in the input space of the black--box model. The generator receives random noise, the black--box's predictions, and a masked version of $C$ ($\widetilde{C}$) where only one category of a single one--hot encoded feature is still equal to one. The critic gets real samples from the data and synthetic samples from the generator, where $N^*$ is generated via Tanh functions and $C^*$ via Gumbel--Softmax functions. In addition, it also receives the black--box's predictions and $\widetilde{C}$ to check the validity of the sample. The critic is trained via the Wasserstein loss, while the generator also adds a reconstruction loss on $C^*$ and $\widetilde{C}$. Note that if the protected features are not part of the black--box model's input space, we can still assess proxy discrimination by generating these features (as part of $N^*$ or $C^*$) and concatenating them to the real samples (as part of $N$ or $C$) given to the critic. We can use $\widetilde{C}$ to assess cases of intersectional discrimination by generating canonical inputs conditional on fixed values for protected features, such as setting ``Male'' for ``Sex.''}
  \label{fig:overview}
\end{figure}

We present \textbf{LUCID--GAN}, which generates canonical inputs via a conditional generative model instead of gradient--based inverse design. LUCID--GAN generates canonical inputs conditional on the predictions of the model under fairness evaluation (see Fig. \ref{fig:overview}a). Using a conditional generative model has two clear benefits. First, we only require a set of (test) samples and corresponding predictions from the model under fairness evaluation, making it a model--agnostic approach. Second, LUCID--GAN generates realistic samples as defined by its objective function. Furthermore, the flexibility of LUCID--GAN is twofold. First, we can extend the canonical inputs with protected features that are not part of the input space of the model under fairness evaluation. Second, the categorical (one--hot encoded) features are part of the conditional vector in the generator, and we can thus condition the canonical inputs on specific feature values (e.g., setting ``Male'' for ``sex''). The first flexibility allows us to assess proxy discrimination, and the second to explicitly check for sources of intersectional discrimination.

LUCID--GAN is an input--based fairness evaluation method which takes a somewhat reverse approach to the statistical output--based metrics. Instead of comparing the predictions, we compare the (protected) feature distributions corresponding to large positive and negative predictions. LUCID--GAN is agnostic towards an a priori decision on the definition of fairness and choice of protected features and instead provides results that suggest potential sources of discrimination \cite{mazijn_sets_2022}. One can then examine the resulting canonical sets from multiple points of view, which are often case--dependent and policy--related \cite{goethals_2022_precof}. By learning the joint conditional distribution of the features on the model's predictions, LUCID--GAN generates a diverse set of realistic synthetic samples that get positive and negative outputs. Instead of a single number and focusing on a select number of protected features, we get an overview of the overall preferences of the model. We can detect combinations of feature values that often appear together (e.g., ``White,'' ``Male,'' ``Married,'' and ``Husband'' in the UCI Adult data set), which gives us insights into the different channels of potential proxy discrimination \cite{adams_2023_proxy}. 

While LUCID--GAN and output--based metrics may sometimes convey the same results, LUCID--GAN can shed light on the potential sources for the statistical disparities. This is crucial as enforcing statistical parity for the wrong reasons can actually harm the protected groups \cite{corbett_measure_2018}. LUCID--GAN reveals a model's preferences for a specific output. In contrast, output--based metrics show the statistical disparities on a benchmark data set that may result from many unknown causes. Overall, we argue that both techniques can be part of the same toolbox as they yield different insights.

We provide a brief overview of the literature on algorithmic fairness, activation maximization, and generative models for tabular data in Section \ref{sec:literature}. In Section \ref{sec:LUCIDGAN}, we present LUCID--GAN and discuss how it solves LUCID's shortcomings. In Section \ref{sec:results}, we show how to generate canonical sets via LUCID--GAN for various fairness evaluations, including direct, proxy, and intersectional discrimination, on the UCI Adult \cite{dua_adult_2017} and COMPAS \cite{angwin_2016_compas} data sets. We find that LUCID--GAN is a valuable addition to the toolbox of algorithmic fairness evaluation, as it offers additional transparency into the sources of discrimination and is agnostic towards an a priori decision on the definition of fairness and choice of protected features.

\section{Background and Related Work}
\label{sec:literature}
The development of LUCID--GAN to expose unethical biases in a model's internal logic connects ideas from the literature on algorithmic fairness, activation maximization using generative models, and recent advancements in GANs (Generative Adversarial Networks) for tabular data. We discuss these different fields below.

\subsection{Algorithmic Fairness}
The inherent ambiguity of viewing the concept of fairness spurred the development of many fairness notions, with over 19 widely accepted definitions \cite{makhlouf_fairness_2021}. Most group fairness notions focus on the equality of outcome by computing statistical parity metrics on a model's output. The two most prominent examples of these statistical output--based metrics are Demographic Parity (DP) and Equality Of Opportunity (EOP) \cite{barocas_fairml_2019}. In DP, we compare the Positivity Rate (PR) of the subpopulations under fairness evaluation, and in EOP, we compare the True Positive Rate (TPR). The choice between DP and EOP entirely depends on the underlying assumptions and worldview of the evaluator \cite{friedler_possibility_2016}. Indeed, even among those two most widely used metrics, substantial philosophical disagreement exists on which one is genuinely fair in each specific context \cite{binns_fairness_2018}. Moreover, they are incompatible, except for highly constrained special cases \cite{kleinberg_inherent_2016}, and it has been empirically shown that inherent trade--offs exist in many practical situations \cite{mazijn_score_2021}. 

In addition, these statistical output--based metrics may suffer from sampling bias and variance as they depend on a benchmark data set \cite{lum_2022_debias}. By reducing the fairness evaluation to a single number, it is hard to verify the validity of the results and to understand exactly why the model is unfair. To enhance the transparency of the statistical parity metrics, there is a strong interaction between output--based fairness evaluations and interpretability methods \cite{meng_mimic_2022}. However, there are many trade--offs and the interpretability methods themselves may suffer from unethical biases \cite{balagopalan_fairness_2022,kleinberg_simple_2019,slack_fooling_2020}. Finally, there is also the selection of protected attributes, which may lead to missing cases of intersectional discrimination, also known as fairness gerrymandering \cite{kearns_2018_gerrymandering}. We argue that LUCID--GAN counters many of these shortcomings by offering additional transparency into the sources of discrimination and being agnostic towards an a priori decision on the definition of fairness and choice of protected features.

\subsection{Activation Maximization}
Performing gradient descent on the input space generates canonical inputs that maximize a specific output activation \cite{simonyan_deep_2014}. However, this gradient--based inverse design approach often leads to unrealistic inputs that obtain a high--confidence score on a specific class. These so--called fool inputs result from discriminative models allocating large areas of high confidence, often much larger than the area occupied by training samples for that class \cite{nguyen_deep_2015}. To avoid the generation of fool inputs, Nguyen et al. \cite{nguyen_plug_2017,nguyen_synthesizing_2016} perform gradient descent in the latent space of a generator network to maximize the output activation in a separate discriminative model. Finally, Odena et al. \cite{odena_conditional_2017} and Zhou et al. \cite{zhou_activation_2018} train an Auxiliary Classifier GAN (AC--GAN) and Activation Maximization GAN (AM--GAN) to maximize the activation of its label in an auxiliary classifier. 

Two important distinctions exist with our application of LUCID--GAN on fairness and tabular data. First, our goal is to generate realistic canonical inputs for any black--box model's predictions, not to stabilize the training of the GAN. Moreover, note that the AC-- and AM--GAN require access to a differentiable model, which we avoid using a Conditional GAN (CGAN) \cite{mirza_conditional_2014}. Second, most previous methods focus on images, where the resulting canonical inputs are individually interpretable and difficult to aggregate \cite{lang_explaining_2021}. In contrast, we infer the overall fairness of a model from the feature distributions by generating a set of canonical inputs.

\subsection{Tabular GAN}
The use of GANs (Generative Adversarial Networks) \cite{goodfellow_gan_2014} for tabular data concerns specific challenges, such as mixed data types, non--Gaussian and multimodal distributions, and sparse and highly imbalanced categorical features \cite{borisov_deep_2021}. To deal with mixed data types, i.e., numerical and one--hot encoded features, different types of output activation functions are used in the generator. For the numerical features, the hyperbolic tangent function (Tanh) is combined with a min--max normalization as a pre--processing step. However, this does not address the non--Gaussian and multimodal distributions. Therefore, Xu et al. \cite{xu_modeling_2019} propose a Conditional Tabular GAN (CTGAN) with mode--specific (min--max) normalization by using a variational Gaussian mixture model. For the categorical features, the Gumbel--Softmax function is used to make the softmax operation in the generator differentiable \cite{jang_gumbel_2017}. To handle the sparse and highly imbalanced categorical features, Xu et al. \cite{xu_modeling_2019} provide a conditional vector to the generator in the form of a masked version of the one--hot encoded features where only one category of a single one--hot encoded feature is still equal to one, and apply a reconstruction loss to the synthetic sample and the masked vector of the selected one--hot encoded feature. They further propose a training--by--sampling technique where the masked one--hot encoded feature is randomly sampled, and its category is sampled from the empirical log--frequency distribution. 

The masked vector can be used to control the generation process similarly to the CGAN framework \cite{mirza_conditional_2014}. For example, \cite{engelmann_conditional_2021} use the masked vector to oversample specific categories in an imbalanced learning problem. Zhao et al. \cite{zhao_ctab-gan_2021} also add the components of the variational Gaussian mixture model to the masked vector to accommodate features with strict upper and lower bounds, e.g., an income of exactly zero. We continue the work on CTGAN by extending the conditional vector of the generator in LUCID--GAN with the black--box model's predictions, which are continuous values that are not part of the synthetic sample, and examining its interaction with the masked vector.

\section{LUCID--GAN}
\label{sec:LUCIDGAN}
We present LUCID--GAN to expose unethical biases in a model’s internal logic by generating canonical inputs. First, we discuss LUCID and the canonical sets as proposed by Mazijn et al. \cite{mazijn_sets_2022}. Then, we introduce LUCID--GAN and show how to generate canonical sets.

\subsection{LUCID}
Shortcomings in the existing output--based group fairness metrics have spurred the research on complementary fairness evaluation methods, which offer additional transparency into the sources of discrimination and are agnostic towards an a priori decision on the definition of fairness and choice of protected features, as this is often case--dependent and policy--related \cite{goethals_2022_precof}. In this spirit, Mazijn et al. \cite{mazijn_sets_2022} propose LUCID, and introduce the notion of a canonical set that allows to evaluate the fairness of a model's decision-making processes. Through gradient-based inverse design, LUCID generates canonical inputs, which can be considered the desired input given a preferred output for a trained model. By repeatedly generating canonical inputs, the resulting canonical set reveals which feature values are essential to obtain specific outputs. This allows for exposing potential unethical biases in the model's internal logic by inspecting the distribution of the protected features. In contrast to output metrics, there is no need for a specific fairness metric, a ground truth, or a benchmark data set.

Following Mazijn et al. \cite{mazijn_sets_2022}, we use LUCID to generate a canonical set for a trained binary classifier.\footnote{We use the same implementation and default values for the hyperparameters as LUCID. For more details see Mazijn et al. \cite{mazijn_sets_2022}. The code for LUCID is available at: \url{https://github.com/Integrated-Intelligence-Lab/canonical\textunderscore sets}.} First, we draw an extensive set of randomly initialized input samples from a uniform distribution. Then, we transform these random input samples into canonical inputs through gradient--based inverse design. Each subsequent transformation results from minimizing the (cross--entropy) loss between the model prediction and the preferred output (e.g., a loan is granted) until the model's prediction is close to its maximum (a predicted probability of 1 in this case). Note that we keep the model parameters fixed throughout the entire procedure. Finally, we inspect the distribution of each protected feature within the canonical set and compare it to the initial random distribution. Several design considerations impact the resulting canonical set, such as the initialization, choice of hyperparameters, and pre-- and post--processing of categorical features. Besides these practical considerations, LUCID has some critical shortcomings, such as the requirement of differentiable models, unrealistic canonical inputs, and difficulties in assessing proxy and intersectional discrimination, which we solve by introducing LUCID--GAN.

\subsection{Specifications of LUCID--GAN}
LUCID--GAN builds upon the CTGAN \cite{xu_modeling_2019} framework. In the class of GANs, a generative model is trained through an adversarial process with a critic model (see Fig. \ref{fig:overview}a). The former aims to create synthetic samples from random noise that fool the latter into judging them as real. Besides the random noise, we provide a conditional vector to the generator, which allows us to control the generation process to some extent \cite{mirza_conditional_2014}. The CTGAN uses a Wasserstein GAN with gradient penalty (WGAN--GP) to prevent common problems in training GAN models, such as mode--dropping, vanishing gradients, and non--convergence \cite{arjovsky_towards_2017,arjovsky_wasserstein_2017,gulrajani_improvedwgan_2017}. Xu et al. \cite{xu_modeling_2019} further propose stabilizing training by following PacGAN \cite{lin_pacgan_2018} and packaging multiple samples together in the critic. Both the generator and the critic consist of fully--connected hidden layers. In the generator, we use batch normalization, ReLU functions, and residual connections on each hidden layer. In the critic, we use leaky ReLU functions and dropout.\footnote{We use the same implementation and default values for the hyperparameters as CTGAN. For more details see Xu et al. \cite{xu_modeling_2019}. The code for CTGAN is available at: \url{https://github.com/sdv-dev/CTGAN}.}

\subsection{Training LUCID--GAN}
The generator in LUCID--GAN generates canonical inputs conditional on the predictions of the black--box under fairness evaluation (see Fig. \ref{fig:overview}b). The black--box's input samples may consist of both numerical ($N$) and one--hot encoded ($C$) features. Both $N$ and $C$ may contain protected features that are not necessarily in the black--box's input space for models which do not directly discriminate against specific features. The numerical features $N$ are pre--processed via mode--specific min--max normalization using a variational Gaussian mixture model, and the categorical features $C$ are one--hot encoded. Note that the processing of the samples for the training of the black--box model and LUCID--GAN does not need to be similar. For example, many tree--based models do not require the one--hot encoding of categorical features.

The generator receives noise which is drawn from a standard normal distribution, and a conditional vector containing the black--box's predictions and a masked version of $C$ ($\widetilde{C}$) where only one category of a single one--hot encoded feature is equal to one. The critic gets real samples from the data and synthetic samples from the generator, where $N^*$ is generated via Tanh functions and $C^*$ via Gumbel--Softmax functions.\footnote{The generator also outputs the components of the variational Gaussian mixture model for each numerical feature as a one--hot encoded vector. The components are part of the input for the critic to address the non--Gaussian and multimodal distributions. They are further used to reverse the mode--specific min--max normalization.} In addition, it also receives the black--box's predictions and $\widetilde{C}$ to check the validity of the sample.

We use $\widetilde{C}$ to handle the sparse and highly imbalanced categorical features via a training--by--sampling technique. It is constructed by randomly sampling a one--hot encoded feature, and subsequently sampling a category from the empirical log--frequency distribution.\footnote{For example, let $C_1 = [a, b]$ and $C_2 = [c, d, e]$ be two one--hot encoded features with a possible empirical sample $C = [C_1, C_2] = [0,1,0,0,1]$, wherein the first one--hot encoded feature we observe the second category $b$, and in the second one--hot encoded feature the third category $e$. We construct a masked version of $C$ by first randomly selecting either $C_1$ or $C_2$, and then by sampling a category from their respective empirical log--frequency distribution. A potentially masked version of $C$ could be $\tilde{C} = [1,0,0,0,0]$ where the first category $a$ of the first one--hot encoded feature $C_1$ is sampled.} All the others values are set to zero. To enforce the generator to generate the sampled category in $\widetilde{C}$, we apply a reconstruction loss to $C^*$ and $\widetilde{C}$. To ensure the validity of the conditional vector, we pick the prediction corresponding to the sample from the empirical distribution. After training, we can use $\widetilde{C}$ to generate canonical inputs conditional on fixed values for a single category, such as setting ``Male'' for ``Sex.''

The critic is trained via the Wasserstein loss, while the generator also adds the reconstruction loss. The reconstruction loss is the cross--entropy between the sampled one--hot encoded feature and its synthetic counterpart in $C^*$. The cross--entropy is only computed on a single one--hot encoded feature, which forces the generator to replicate this condition in the synthetic sample.

\subsection{Generating Canonical Sets with LUCID--GAN}
After training, we use the generator of LUCID--GAN to generate canonical inputs (see Fig. \ref{fig:overview}c). Similar to the CGAN framework \cite{mirza_conditional_2014}, we can use the conditional vector to control the generation process. Indeed, setting the prediction in the conditional vector to a specific value corresponds to maximizing the output activation via gradient descent. For example, in the case of a binary classifier, we can generate a canonical set that reveals the preferred output for receiving a ``positive'' and ``negative'' decision by setting the prediction in the conditional vector equal to a predicted probability of 1 and 0, respectively. Instead of comparing the predictions, we compare the (protected) feature distributions corresponding to large positive and negative predictions.

Using a generative model has the benefits of working for any black--box model and generating realistic synthetic samples. Note that if the protected features are not part of the black--box model's input space, we can still assess proxy discrimination by generating these features (as part of $N^*$ or $C^*$) and concatenating them to the real samples (as part of $N$ or $C$) given to the critic. We use $\widetilde{C}$ to assess intersectional discrimination by generating canonical inputs conditional on fixed values for protected features, such as setting ``Male'' for ``Sex.'' This allows us to compare feature distributions for many possible scenarios where otherwise, data would be scarce, and the estimates of output--based metrics would be unreliable \cite{lum_2022_debias}.

\section{Experiments and Discussion}
\label{sec:results}
We show how to generate canonical sets via LUCID--GAN for various fairness evaluations, including direct, proxy, and intersectional discrimination, on the UCI Adult \cite{dua_adult_2017} and COMPAS \cite{angwin_2016_compas} data sets. In the UCI Adult data set, we predict if a person earns more or less than $\$50,000$ per year, with more being the preferred output. For COMPAS, the task is to predict if a person will commit recidivism in the next two years, with no recidivism being the preferred output as the person can be released on bail. For both data sets, we consider ``Race'' and ``Sex'' to be the (legally) protected features, and for UCI Adult we also consider ``Marital Status'' and ``Relationship.'' The UCI Adult data set has a fixed test set, while for the COMPAS data set, we sample $20\%$ from the training data as the test set. The samples and predictions of the test set are also used to train LUCID--GAN and to compute the statistical output--based metrics DP and EOP.

We compare LUCID--GAN with LUCID and the statistical output--based metrics DP and EOP (by comparing the TP and TPR, respectively) for the case of direct discrimination (see Fig. \ref{fig:directdiscrimination} and Table \ref{tab:output}). We further demonstrate how to apply LUCID--GAN to the cases of proxy (See Fig. \ref{fig:indirectdiscrimination}) and intersectional (see Fig. \ref{fig:intersectional}) discrimination. As LUCID only works for differentiable models and DP and EOP are often used for binary classification tasks, we use binary fully-connected neural network classifiers. For LUCID--GAN, the extension to other black--box classifiers is trivial, as it only requires samples and their corresponding predictions, making it a model--agnostic approach.

The classifiers consist of hidden layers with ReLU activation functions and a softmax output layer with two output nodes. The number of hidden layers and nodes is decided by the accuracy on a validation set ($20\%$ from the training set) which is $83.9\%$ for UCI Adult and $64.0\%$ for COMPAS. This performance is in line with the standard benchmarks. We choose this standard architecture as the point of this experiment is not to achieve state--of--the--art performance but to demonstrate the capabilities of LUCID--GAN, which does not depend on the quality of the underlying model. Note that the computational complexity of LUCID--GAN lies in the training of the generator and the critic, which for our experiments was only a matter of minutes on a consumer CPU. After training, the generation of canonical inputs for various fairness evaluations is done via single forward passes through the generator.

\subsection{Direct Discrimination}
\begin{figure}[H]
  \centering
  
  \begin{subfigure}[b]{1\textwidth}
  \caption{UCI Adult}
  \centering 
  \includegraphics[width=0.90\textwidth]{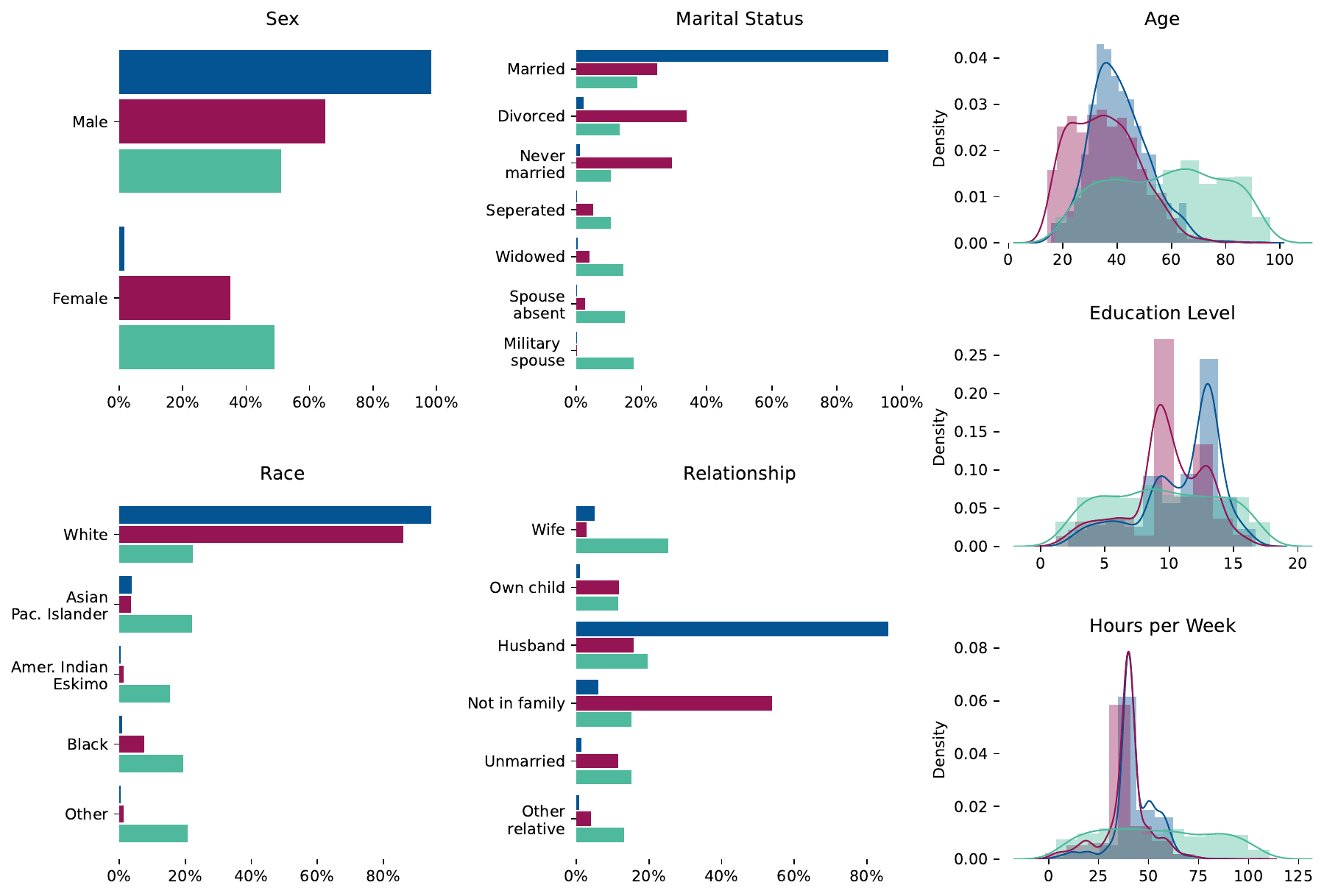}
  \label{fig:directdiscrimination_adult}
  \end{subfigure}
  
  \begin{subfigure}[b]{1\textwidth}
  \centering
  \caption{COMPAS}
  \includegraphics[width=0.90\textwidth]{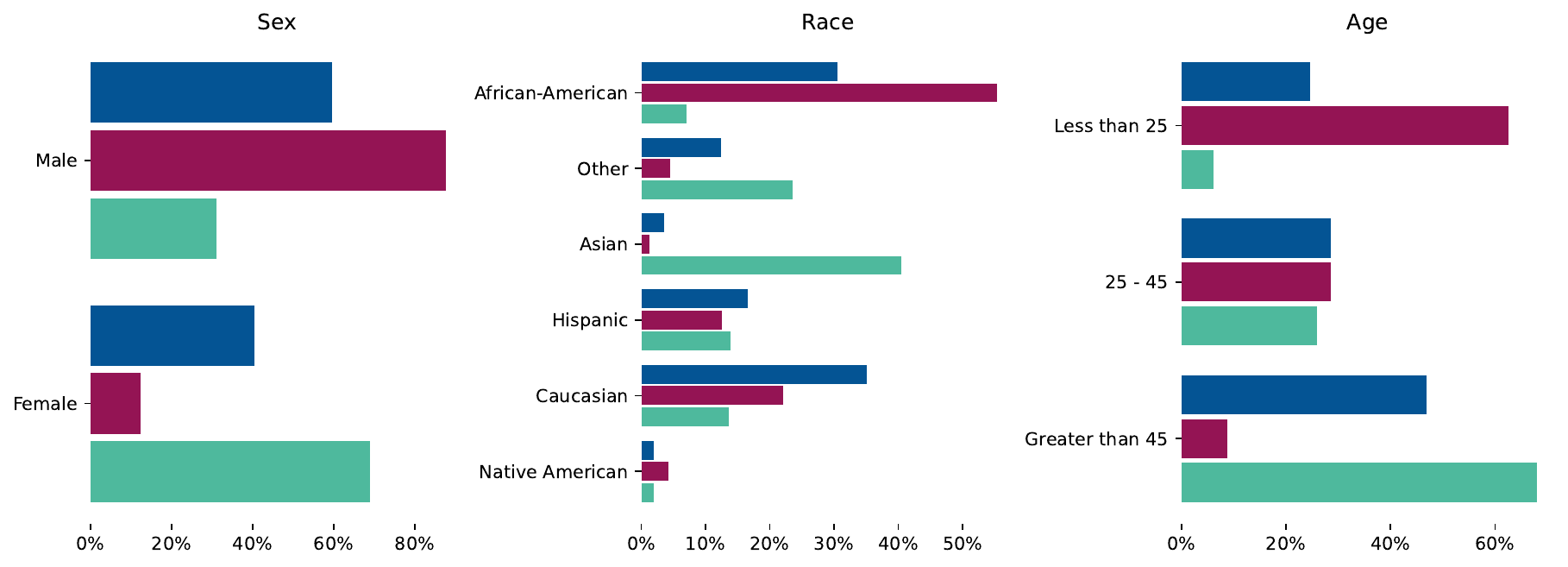}
  \label{fig:directdiscrimination_compas}
  \end{subfigure}
  
  \caption{Locating direct discrimination. The feature distributions for 1000 synthetic samples are shown for the positive and negative outputs in LUCID--GAN (dark blue and dark red, respectively) and positive outputs in LUCID (light green). For the UCI Adult data set, we show the protected features ``Sex,'' ``Race,'' ``Marital Status'' and ``Relationship.'' We further show the features ``Age,'' ``Education Level,'' and ``Hours per Week.'' For the COMPAS data set, we show the protected features ``Sex'' and ``Race.'' We further show ``Age,'' which is a categorical feature. Note that to interpret the results of LUCID--GAN, we need to compare the positive and negative canonical sets, while for LUCID we can compare the positive canonical set with an initial random uniform distribution (see also Appendix A). Only inspecting the feature distributions of either the positive or negative canonical set of LUCID--GAN can lead to misleading results as the generator is trained to mimic the underlying distribution of its training samples.}
  \label{fig:directdiscrimination}
\end{figure}

We compare LUCID--GAN with LUCID and the statistical output--based metrics DP and EOP (by comparing the disparities between TP and TPR, respectively) for the case of direct discrimination (see Fig. \ref{fig:directdiscrimination} and Table \ref{tab:output}). After the classifiers are trained, we train LUCID--GAN on the test set and then generate 1000 synthetic samples for a positive and negative output (i.e., a predicted probability of 1 and 0, respectively). We refer to them as the positive and negative canonical sets. Note that we require no access to the training data and treat the underlying classifier as a black--box. For LUCID, we also generate 1000 synthetic samples for a positive output starting from an initial random uniform distribution (see also Appendix A). Finally, we compute the TP and TPR of all the subpopulations for the protected features. A disparity between these metrics indicates potential discrimination towards the group with lower values.

We locate unfairness in the model's decision--making process with LUCID--GAN by comparing the feature distributions of the positive and negative canonical sets. Only inspecting the feature distributions of either the positive or negative canonical set can lead to misleading results, as the generator is trained to mimic the underlying distribution of its training samples. For example, in both the UCI Adult and COMPAS data sets, more samples correspond to the ``Male'' category than the ``Female'' category for the feature ``Sex'' which results in more males in both the positive and negative canonical sets.

While the results in our experiments are quite clear on visual inspection and the data sets are not high--dimensional, this may not always be the case, and presenting the results as such can be unfeasible and ambiguous. In that case, we suggest using distance metrics, such as the Wasserstein or Jensen--Shannon distance. However, computing these metrics yields no additional insights into our experiments.

In Fig. \ref{fig:directdiscrimination}, we see that LUCID--GAN generates more realistic canonical inputs than LUCID. The most prominent examples are the continuous features, where LUCID often generates unrealistic values while LUCID--GAN generates feature distributions that closely match the expected empirical distributions. For example, in the feature ``Education Level'' of the UCI Adult data set, we see a spike in high--school graduates (9 years), bachelors (13 years), and masters (14 years). The realism is a result of the adversarial process where the critic ensures that the generator outputs a diverse set of synthetic samples which look similar to the training samples. The realism also seems to lead to more outspokenness in LUCID--GAN's canonical inputs for at least two reasons. First, the synthetic samples must respect all the dependencies between the features. For example, the combination of ``Male,'' ``Husband,'' and ``Married'' is a strongly preferred sample in LUCID--GAN's positive canonical set. The strong co--occurrence of these feature values may indicate a potential for proxy discrimination. Second, it may be easier for LUCID to change the continuous features than the categories, as a small change in categorical features does not make any difference after post--processing. Any change in the continuous features remains, while the categories need to shift from one category to another due to the one--hot encoding.

\begin{landscape}
\begin{table}[t!]
\caption{Positivity Rate (PR) and True Positive Rate (TPR) of the subpopulations for the protected features in UCI Adult and COMPAS.}
\label{tab:output}
\begin{center}
\begin{small}
\begin{sc}
\begin{tabular}{c|cccccccc}
    \toprule
\small UCI Adult \\
    &&&&&&&\\
     \emph{Sex}  & Male & Female \\
%    \midrule
    PR & 31.0 & 11.3 \\
    TPR & 73.0 & 72.3 \\
%  \midrule
    &&&&&&&\\
     \emph{Race} & White & \shortstack{Asian Pac.\\Islander} & \shortstack{Amer. Indian\\Eskimo}&Black&Other\\
    PR & 26.0 & 29.7 & 12.8 & 11.9 & 19.7\\
    TPR & 73.2 & 72.8 & 66.7 & 64.9 & 77.8\\
    &&&&&&&\\
     \emph{Relationship} &Wife & Own child&Husband &Not in family& Other relatives & Unmarried\\
%    \midrule
    PR & 47.0 & 1.9 & 45.6 & 10.2 & 3.3 & 5.6 \\
    TPR & 71.0 & 100.0 & 72.8 & 78.7 & 100.0 & 81.8 \\
        &&&&&&&\\
      \emph{Marital Status} &Married & Divorced & \shortstack{Never\\married} & Separated & Widowed & \shortstack{Spouse\\absent} & \shortstack{Military\\spouse}\\
%    \midrule
    PR & 45.3 & 9.7 & 4.7 & 7.0 & 9.1 & 12.6 & 36.4\\
    TPR & 72.6 & 75.0 & 86.2 & 55.6 & 85.7 & 100.0 & 80.0\\\\
%    \midrule
     \midrule
    \small COMPAS \\
        &&&&&&&\\
     \emph{Sex}   & Male & Female \\
    PR & 53.6 & 62.7 \\
    TPR & 39.4 & 38.9 \\
        &&&&&&&\\
\emph{Race} & \shortstack{African\\ American} & Asian & Caucasian & Hispanic &\shortstack{Native \\American}&Other\\
PR & 49.2 & 71.4 & 61.9 & 62.5 & 60.0 & 54.0 \\
TPR & 37.6 & 0.00 & 41.9 & 52.0 & 60.0 & 25.0 \\
 \bottomrule
\end{tabular}
\end{sc}
\end{small}
\end{center}
\end{table}
\end{landscape}

In Table \ref{tab:output}, we see that the DP and EOP disparities do not always point to the same conclusions. For example, the disparity between the PR of ``Male'' and ``Female'' indicates a violation of DP, while their TPRs indicate that there is EOP for both the UCI Adult and COMPAS data sets. Additionally, many categories, including ``Military Spouse'' in the UCI Adult data set and ``Asian'' in the COMPAS data set, do not contain sufficient data points to make any conclusions \cite{lum_2022_debias}. Finally, the output--based metrics do not give any insights into the actual drivers of the unfairness. There are some notable differences between the output--based metrics and LUCID--GAN. For example, in the UCI Adult data set, the DP metric indicates that ``Wife'' is the preferred group in the ``Relationship'' feature. In contrast, the canonical sets indicate a strong preference for ``Husband.''

\subsection{Proxy Discrimination}
We retrain the models without the protected features ``Race'' and ``Sex'' and obtain a similar accuracy on the test set. We then generate a positive and negative canonical set including the left--out protected features, by generating these features (as part of $C^*$) and concatenating them to the real samples (as part of $C$) given to the critic. LUCID--GAN now receives predictions from the black--box model, which does not have ``Race'' and ``Sex'' in its input space. For the UCI Adult data set, we keep the protected features ``Marital Status'' and ``Relationship'' and assume that there are no protected features left in the COMPAS data set. For the PR and TPR, we obtain similar scores as for direct discrimination and therefore refer to Table \ref{tab:output} for a comparison. This confirms many previous findings that removing the protected attributes does not generally improve the disparity in the statistical output--based metrics \cite{dwork_2012_fairness}.

\begin{figure}[ht!]
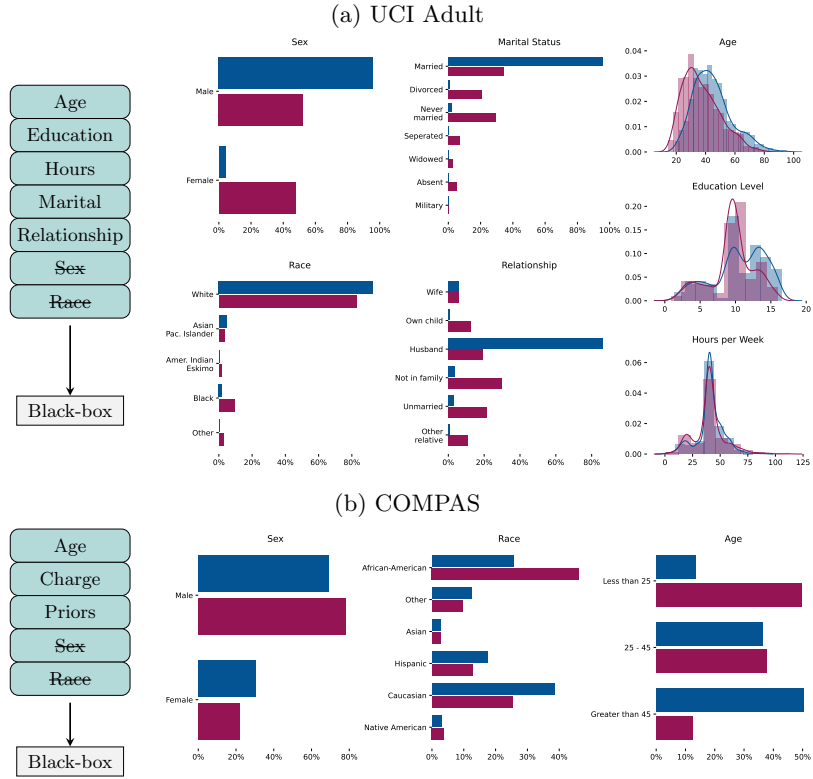

  \centering
  
  \begin{subfigure}[b]{1\textwidth}
  \centering
  \caption{UCI Adult}
  \includestandalone[mode=buildnew, width=0.90\textwidth]{figs/indirect_adult}
  \label{fig:indirectdiscrimination_adult}
  \end{subfigure}
  
  \begin{subfigure}[b]{1\textwidth}
  \centering
  \caption{COMPAS}
  \includestandalone[mode=buildnew, width=0.90\textwidth]{figs/indirect_compas}
  \label{fig:indirectdiscrimination_compas}
  \end{subfigure}
  
  \caption{Locating proxy discrimination. The feature distributions for 1000 synthetic samples are shown for the positive and negative outputs in LUCID--GAN (dark blue and dark red, respectively). For the UCI Adult data set, we show the protected features ``Sex,'' ``Race,'' ``Marital Status'' and ``Relationship.'' We further show the features ``Age,'' ``Education Level,'' and ``Hours per Week.'' For the COMPAS data set, we show the protected features ``Sex'' and ``Race.'' We further show the feature ``Age.'' Note that to interpret the results of LUCID--GAN, we need to compare the positive and negative canonical sets. Only inspecting the feature distributions of either the positive or negative canonical set of LUCID--GAN can lead to misleading results as the generator is trained to mimic the underlying distribution of its training samples.}
  \label{fig:indirectdiscrimination}
\end{figure}

By comparing the canonical sets from the case of proxy discrimination (see Fig. \ref{fig:indirectdiscrimination}) with those of direct discrimination (see Fig. \ref{fig:directdiscrimination}), we see that there are some notable differences (see also Appendix B). For example, in the COMPAS data set, the disparity between ``Male'' and ``Female'' in the positive and negative canonical sets has almost entirely disappeared. At the same time, the relative values in the ``Race'' feature remain mostly unchanged. This may indicate that the previous black--box model was directly discriminating based on ``Sex,'' while the discrimination against ``Race'' is a combination of both direct and proxy discrimination. For the UCI Adult data set, we find that the discrimination against ``Sex'' and ``Race'' remains. This may result from the strong dependencies between many features, such as ``White,'' ``Male,'' ``Married,'' and ``Husband.'' 

\begin{figure}[t!]
  \centering
 
  \begin{subfigure}[b]{1\textwidth}
  \caption{UCI Adult}
  \centering 
  \includegraphics[width=0.90\textwidth]{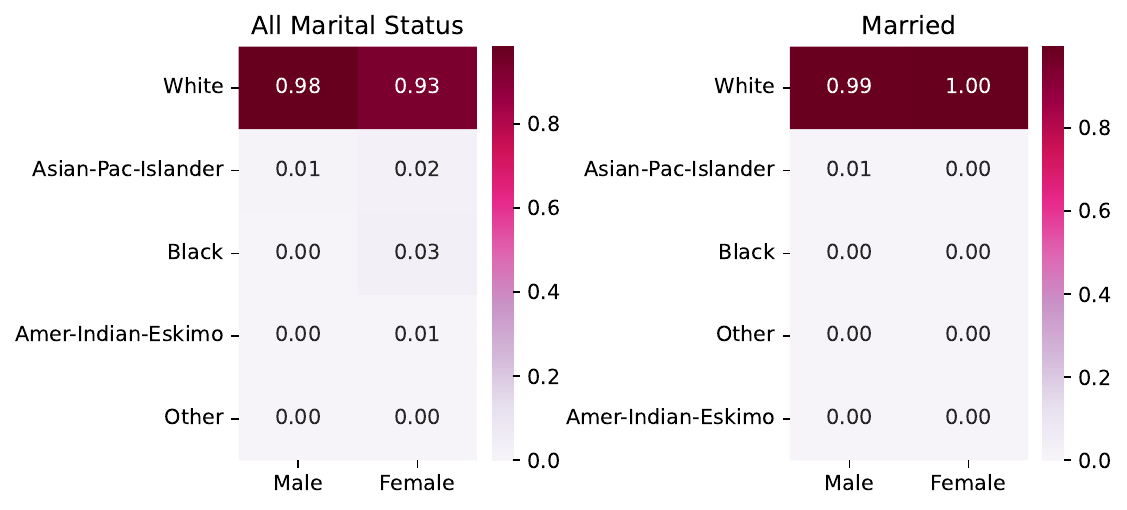}
  \label{fig:intersectional_adult}
  \end{subfigure}
  
  \begin{subfigure}[b]{1\textwidth}
  \centering
  \caption{COMPAS}
  \includegraphics[width=0.90\textwidth]{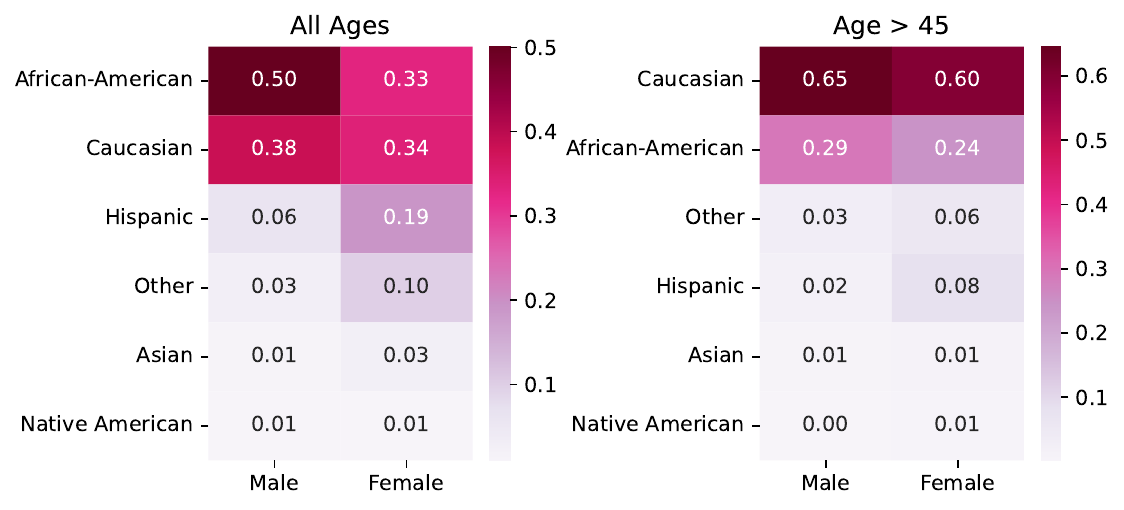}
  \label{fig:intersectional_compas}
  \end{subfigure}
  
  \caption{Locating intersectional discrimination. The percentage--wise amount of men and women per ``Race'' category which receive a positive output is shown for 1000 synthetic samples where ``Sex'' is fixed as ``Male'' and ``Female,'' respectively. Additionally, for the UCI Adult data set, we show the percentage--wise amount of men and women per ``Race'' category which receive a positive output when the ``Marital Status'' is fixed as ``Married,'' and for the COMPAS data set, when ``Age'' is fixed as ``$>45$.''}
  \label{fig:intersectional}
\end{figure}

\newpage
\clearpage
\subsection{Intersectional Discrimination}
We use the original models from the direct discrimination evaluation with the protected features ``Race'' and ``Sex'' included. In this intersectional discrimination evaluation, we generate 1000 positive canonical inputs where we keep ``Sex'' fixed as ``Male'' and 1000 positive inputs where we keep ``Sex'' fixed as ``Female'' by using $\widetilde{C}$ in the conditional vector. Additionally, for the UCI Adult data set, we generate 1000 positive canonical inputs where we also keep ``Marital Status'' fixed as ``Married,'' and for the COMPAS data set, we keep ``Age'' fixed as ``$>45$.'' We show in each case percentage--wise the amount of men and women per ``Race'' category which receive a positive output (see Fig. \ref{fig:intersectional}). On the heat maps, we show a two-- and three--dimensional representation of the frequency of protected features at various intersections. For the UCI Adult data set, we find potential discrimination against women conditional on being ``White,'' which entirely disappears if we add the condition of ``Married.'' For the COMPAS data set, we see indications of bias toward women conditional on being ``Caucasian'' and ``African-American.'' Additionally, when conditioning on ``$>45$'' the number of positive outputs is considerably larger for ``Caucasian'' compared to ``Hispanic'' and ``African-American.'' We believe that these heat maps are an ideal method for tracing potential sources of intersectional discrimination, especially when data samples are scarce. 

\section{Conclusion}
\label{sec:conclusion}
The increasing use of Artificial Intelligence (AI) algorithms in (semi--)automated decision--making processes has raised concerns about discriminatory decision patterns. The literature on algorithmic fairness has primarily focused on defining unfairness and eliminating unethical biases, while recent efforts focus on rigorously detecting it. A recent proposal in this direction is LUCID (Locating Unfairness through Canonical Inverse Design), which generates canonical sets by performing gradient descent on the input space, revealing a model's desired input given a preferred output.

We present LUCID--GAN, which generates canonical inputs via a conditional generative model instead of gradient--based inverse design. Using a conditional generative model has several benefits, including that it applies to non--differentiable models, ensures that a canonical set consists of realistic inputs, and allows us to assess proxy and intersectional discrimination. LUCID--GAN is an input--based fairness evaluation method which takes a somewhat reverse approach to the statistical output--based metrics. Instead of comparing the predictions, we compare the (protected) feature distributions corresponding to large positive and negative predictions. The resulting canonical sets contain valuable information about the model’s mechanisms, i.e., which feature values are essential to obtain specific outputs. This allows us to expose potential unethical biases in its internal logic by inspecting the distribution of the protected features.

We show how to generate canonical sets via LUCID--GAN for various fairness evaluations, including direct, proxy, and intersectional discrimination, on the UCI Adult and COMPAS data sets. It allows for rigorously detecting unethical biases in black--box models without requiring access to the training data. Overall, we argue that LUCID--GAN is a valuable addition to the toolbox of algorithmic fairness evaluation, as it offers additional transparency into the sources of discrimination and is agnostic towards an a priori decision on the definition of fairness and choice of protected features.

\clearpage
\bibliographystyle{splncs04}
\bibliography{biblio.bib}

\clearpage
\appendix
\section{LUCID}
\label{sec:applucid}

\begin{figure}[ht!]
  \centering
  
  \begin{subfigure}[b]{1\textwidth}
  \caption{UCI Adult}
  \centering 
  \includegraphics[width=0.90\textwidth]{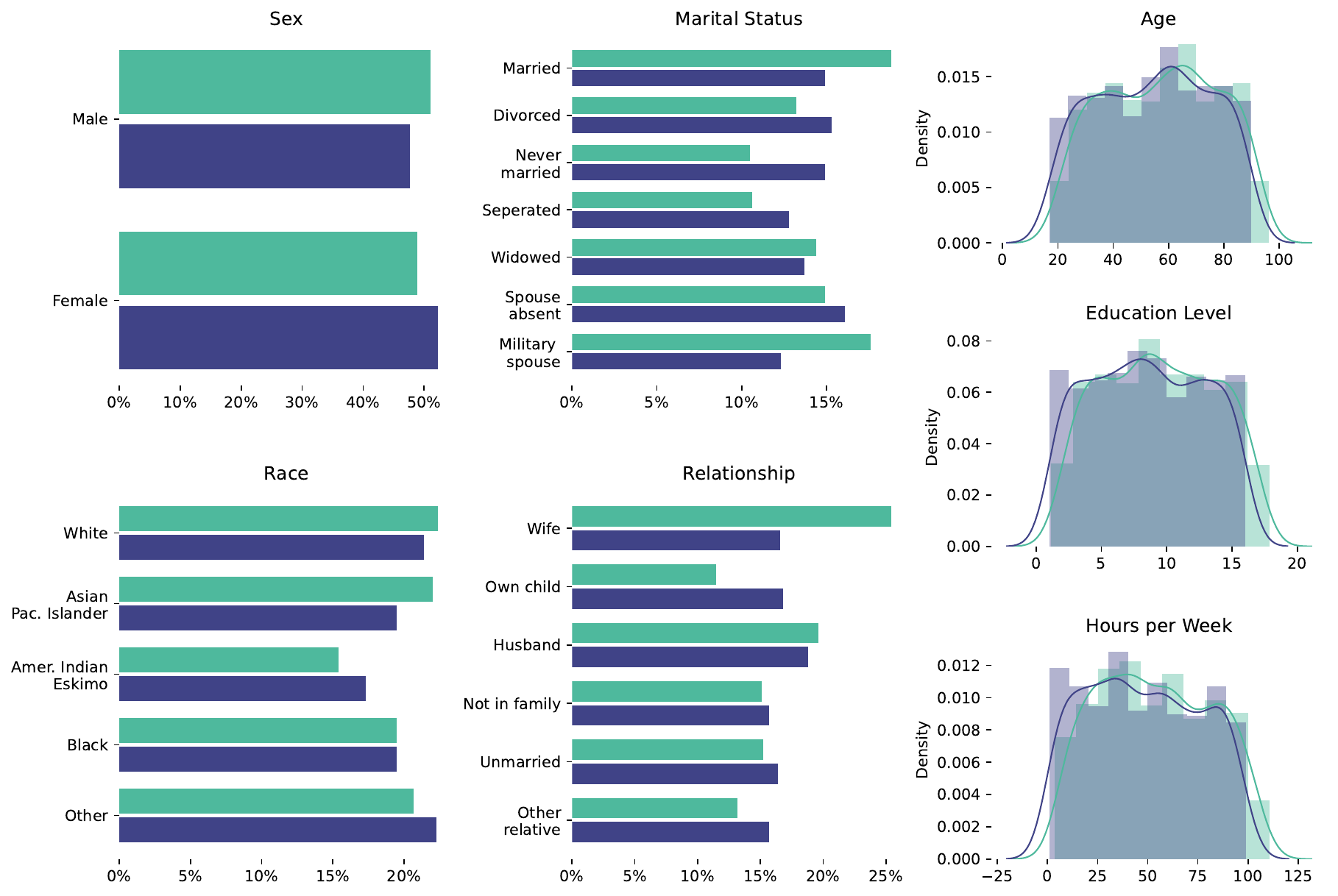}
  \label{fig:lucid_adult}
  \end{subfigure}
  
  \begin{subfigure}[b]{1\textwidth}
  \centering
  \caption{COMPAS}
  \includegraphics[width=0.90\textwidth]{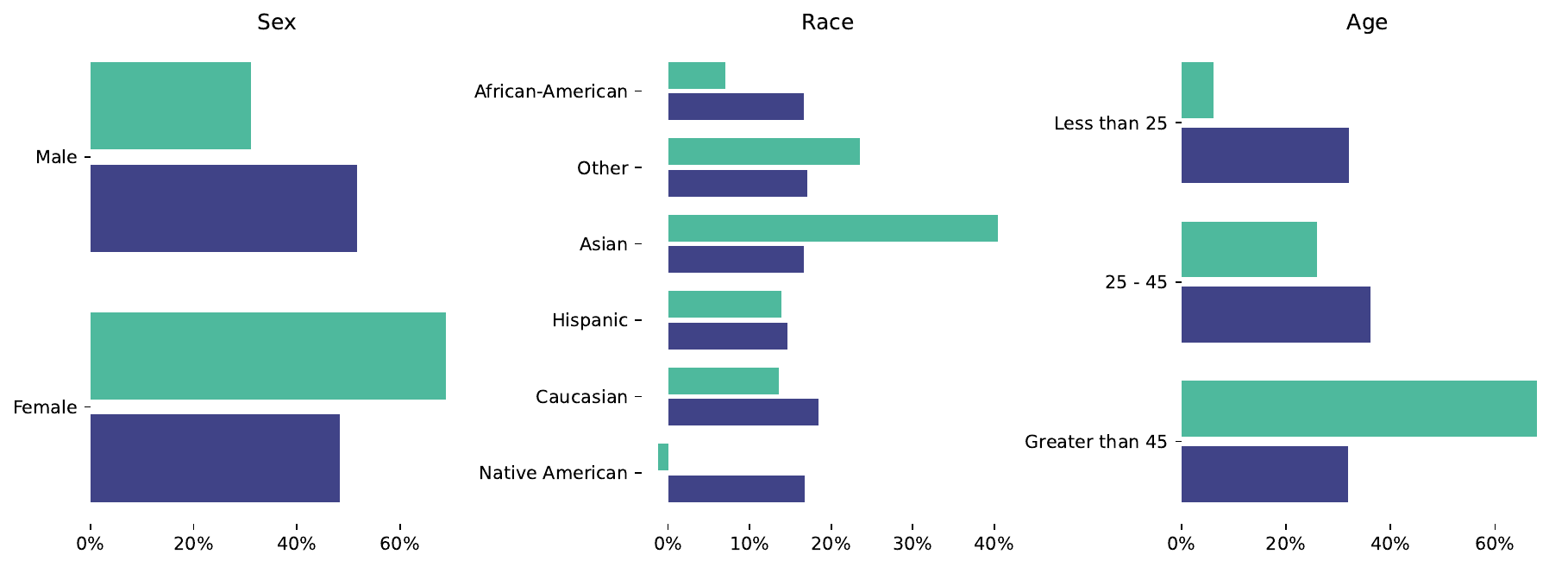}
  \label{fig:lucid_compas}
  \end{subfigure}
  
  \caption{Locating direct discrimination. The feature distributions for 1000 synthetic samples are shown for the positive outputs in LUCID (light green) starting from an initial random uniform distribution (dark blue). For the UCI Adult data set, we show the protected features ``Sex,'' ``Race,'' ``Marital Status,'' and ``Relationship.'' We further show the features ``Age,'' ``Education Level,'' and ``Hours per Week.'' For the COMPAS data set, we show the protected features ``Sex'' and ``Race.'' We further show the feature ``Age.''}
  \label{fig:lucid}
\end{figure}

\clearpage

\section{Comparing Direct and Proxy Discrimination}
\label{sec:appcompare}

\begin{figure}[ht!]
  \centering
  
  \begin{subfigure}[b]{1\textwidth}
  \caption{UCI Adult}
  \centering 
  \includegraphics[width=0.90\textwidth]{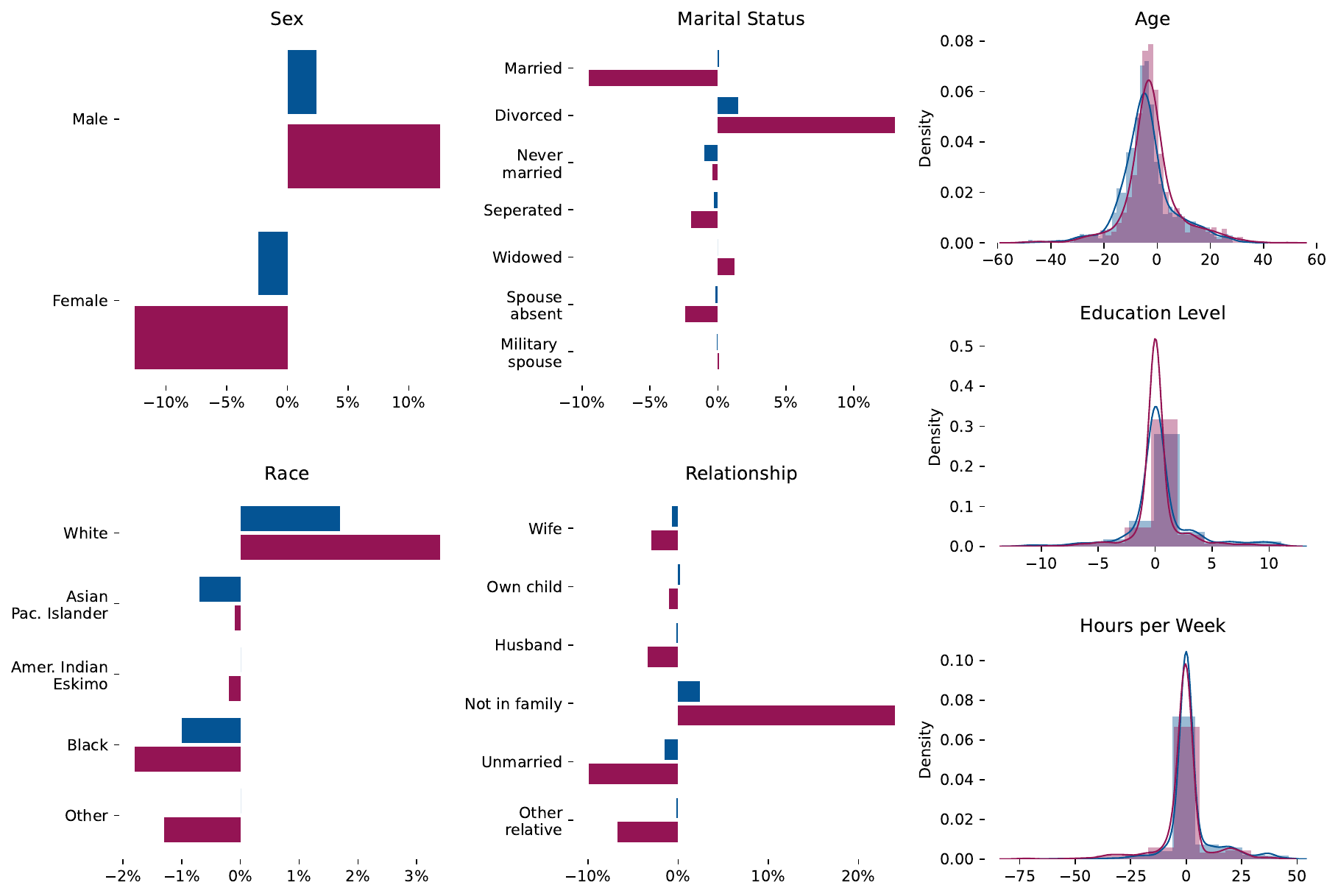}
  \label{fig:compare_adult}
  \end{subfigure}
  
  \begin{subfigure}[b]{1\textwidth}
  \centering
  \caption{COMPAS}
  \includegraphics[width=0.90\textwidth]{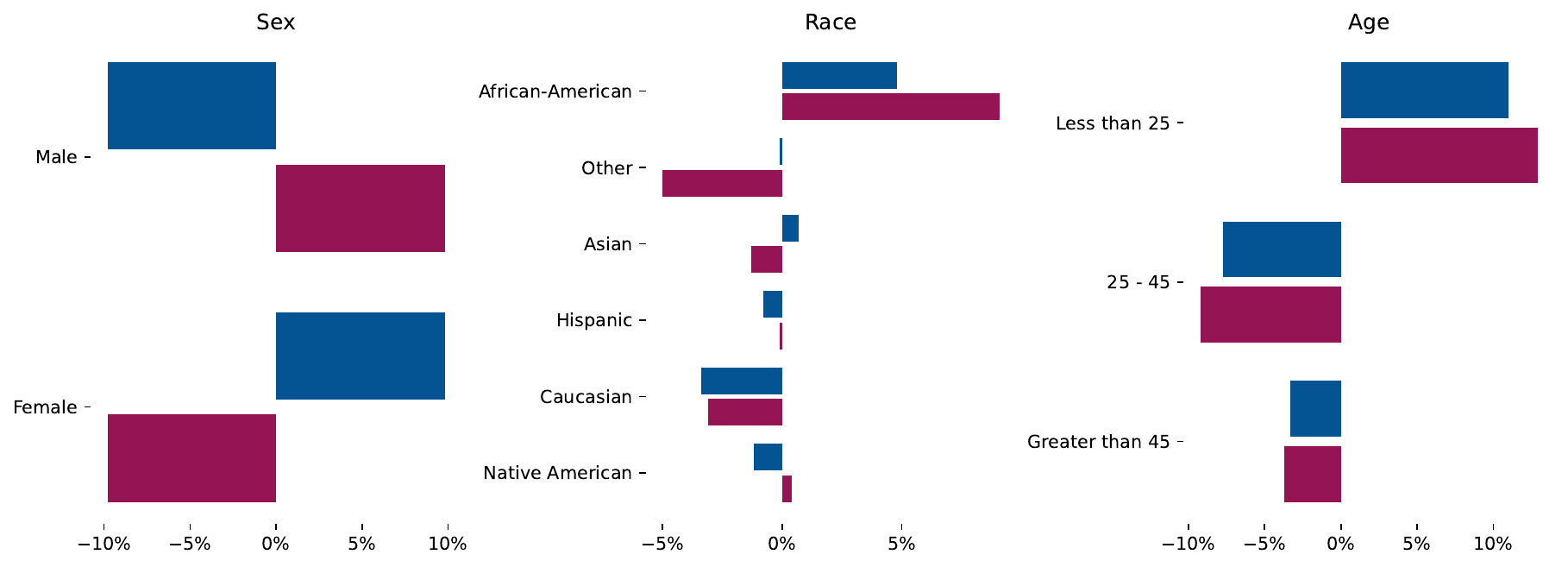}
  \label{fig:compare_compas}
  \end{subfigure}
  
  \caption{The difference between the feature distributions of direct and proxy discrimination for the positive and negative outputs in LUCID--GAN (dark blue and dark red, respectively) are shown. For the UCI Adult data set, we show the protected features ``Sex,'' ``Race,'' ``Marital Status,'' and ``Relationship.'' We further show the features ``Age,'' ``Education Level,'' and ``Hours per Week.'' For the COMPAS data set, we show the protected features ``Sex'' and ``Race.'' We further show the feature ``Age.'' A positive value indicates that the feature appears more frequently in the distribution of direct discrimination.}
  \label{fig:compare}
\end{figure}

\end{document}